\documentclass{article}

\usepackage{spconf,amsmath,graphicx}
\usepackage{amssymb}
\usepackage{amsfonts}
\usepackage{array}
\usepackage{booktabs}
\usepackage{graphicx}
\usepackage{xurl}
\PassOptionsToPackage{hyphens}{url}
\usepackage{color, colortbl}
\definecolor{Gray}{gray}{0.9}

\usepackage[capitalize]{cleveref}
\crefname{section}{Sec.}{Secs.}
\Crefname{section}{Section}{Sections}
\Crefname{table}{Table}{Tables}
\crefname{table}{Tab.}{Tabs.}

\title{Don't Look into the Sun:\\Adversarial Solarization Attacks on Image Classifiers}
\name{Paul Gavrikov$^{1}$\thanks{We thank Julia Schneider for her curiosity which lead to the development of this paper. This project is funded by the German Ministry for Science, Research and Arts, Baden-Wuerttemberg under Grant 32-7545.20/45/1 (Q-AMeLiA).} \qquad Janis Keuper$^{1,2}$\footnotemark[1]\thanks{© 2023 IEEE. Personal use of this material is permitted. Permission from IEEE must be obtained for all other uses, in any current or future media, including reprinting/republishing this material for advertising or promotional purposes, creating new collective works, for resale or redistribution to servers or lists, or reuse of any copyrighted component of this work in other works.}}
\address{$^{1}$IMLA, Offenburg University, $^{2}$Fraunhofer ITWM}

\begin{document}
\maketitle

\begin{abstract}

Assessing the robustness of deep neural networks against out-of-distribution inputs is crucial, especially in safety-critical domains like autonomous driving, but also in safety systems where malicious actors can digitally alter inputs to circumvent safety guards. However, designing effective out-of-distribution tests that encompass all possible scenarios while preserving accurate label information is a challenging task. Existing methodologies often entail a compromise between variety and constraint levels for attacks and sometimes even both. In a first step towards a more holistic robustness evaluation of image classification models, we introduce an attack method based on image solarization that is conceptually straightforward yet avoids jeopardizing the global structure of natural images independent of the intensity. Through comprehensive evaluations of multiple ImageNet models, we demonstrate the attack's capacity to degrade accuracy significantly, provided it is not integrated into the training augmentations. Interestingly, even then, no full immunity to accuracy deterioration is achieved. In other settings, the attack can often be simplified into a black-box attack with model-independent parameters. Defenses against other corruptions do not consistently extend to be effective against our specific attack.

{\noindent\textbf{Project website:} \url{https://github.com/paulgavrikov/adversarial_solarization}}
\end{abstract}
\begin{keywords}
image classification, robustness, adversarial attacks, deep learning
\end{keywords}
\section{Introduction}
 Deep neural networks (DNNs) account for breakthroughs in multiple domains, notably excelling in tasks like image classification. However, the models' exceptional performances on test datasets can be elusive. Despite their remarkable capacity to learn from training data, DNNs often struggle to transcend the boundaries of their training distribution. This limitation can result in unpredictable behavior in the face of distribution shifts, such as changes in lighting, weather, image compression (e.g., JPEG), and more, while human predictions remain unaffected. 
\begin{figure}
    \centering
    \includegraphics[width=\columnwidth]{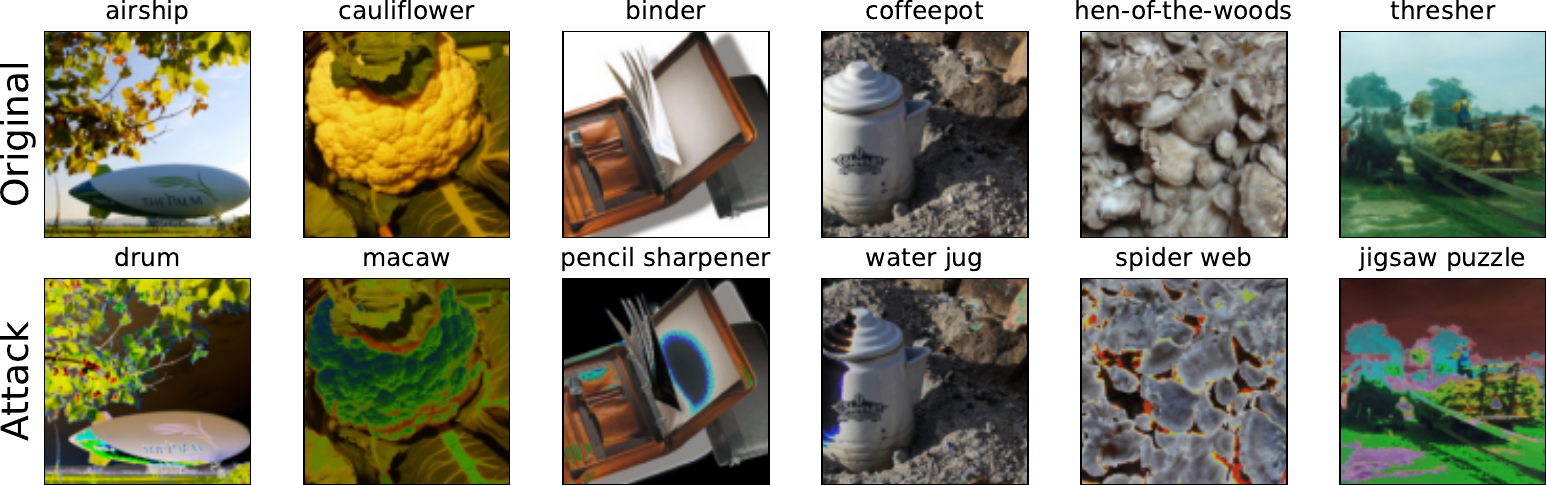}
    \caption{Examples of solarized ImageNet images that flip the previously correct top-1 predictions of ResNet-50 into non-sensical labels. Predicted labels are displayed above the images.}
    \label{fig:solarization_attack}
\end{figure}
Improving the generalization of neural networks to out-of-distribution (OOD) data remains a challenging task, as it necessitates a comprehensive understanding of neural network dynamics that current research has not fully achieved. A critical step towards this objective is the accurate measurement of a neural network's robustness when confronted with OOD data. In an ideal scenario, for a robust model $\mathcal{F}$ parameterized by $\theta$, and for any input-output pair $(x, y)$ belonging to the dataset $\mathcal{D}$, we seek to ensure that:
\begin{equation}
\mathcal{F}(x; \theta) = \mathcal{F}\big(\Psi(x); \theta\big) = y
\end{equation}
Here, $\Psi(\cdot)$ represents a function that modifies input samples. However, characterizing $\Psi$ presents a substantial challenge. Firstly, it should encompass all reasonable modifications that can occur in practice. Secondly, it should adapt the intensity of these modifications without compromising the recognizability of labels by humans.

Previous research has approached this problem in various ways. Adversarial attacks \cite{SzegedyZSBEGF13}, for instance, define $\Psi(x) = x + \delta$, where $\delta$ represents a perturbation. This approach, by definition, covers all modifications, including those that can naturally occur or be artificially introduced through digital manipulation. However, it also introduces the possibility of producing modified images that belong to a different class. To mitigate this, constraints are imposed on $\delta$, typically by ensuring that $||\delta||_p \leq \epsilon$, where $||\cdot||_p$ is an $\ell_p$-norm distance and $\epsilon$ is a small threshold. This constraint ensures that $\Psi(x)$ maintains a high degree of similarity to the original sample $x$ and does not alter the semantic meaning of the image for human observers. However, it limits the scope of modifications to a constrained space, leaving out modifications that exceed these bounds but may still retain the original label.

An alternative approach involves benchmarking \textit{common corruptions} that naturally occur during image processing, such as noise, blurring, variations in weather conditions, or digital artifacts (\textit{ImageNet-C}) \cite{hendrycks2018benchmarking}. Here every corruption $c$ can be characterized by a specific $\Psi_c$  and, thus allows for a more nuanced tuning of the intensity per corruption. Yet, most corruptions will only guarantee label integrity in a specific intensity range(s). For instance, de-/increasing the brightness of an image to the extremes will collapse to a single color and destroy any objects in the image. Additional problems arise due to non-convexity. Weather effects like snow may hide the objects depending on the placement of the snow particles. \textit{Hendrycks et al.} avoided this problem by selecting corruptions intensities that fool a simple model like \textit{AlexNet} \cite{NIPS2012_c399862d} with 100\% success rate. While this may be an acceptable way to compare models to other models it does not guarantee label integrity to human observers. Since models and humans fundamentally differ in their methods of recognition (e.g., \cite{geirhos2018imagenettrained}), robustness on common corruptions doesn't necessary translate to human robustness. Contrary, assessment with human observers require a high level of carefulness and are, thus, costly. Scaling these experiments to the size of \textit{ImageNet} is also not realistic. Hence, previous human trials were limited to small subsets with coarse labels, e.g., \cite{geirhos2021partial}.

As some of the corruptions are computationally expensive, benchmarks relying on them are often distributed in the form of datasets. The finite nature of these datasets only reflects a discrete sampling of intensities and cannot detect defenses overfitting to those.
This may explain why improvements on \textit{ImageNet-C} do not always transfer to other (similar) corruption benchmarks \cite{mintun2021on}.

This paper combines the strenghts of adversarial attacks and common corruptions and introduces a new adaptive adversarial attack following the concept of \textit{image solarization}. The proposed attack strategically fools state-of-the-art image classification models while maintaining a critical guarantee: the integrity of labels in the face of corruption, independent of the severity. Solarization, an intriguing photographic effect, arises when excessively bright objects exhibit a tone reversal due to overexposure, evoking the appearance of gray or black objects. 
Although modern cameras are no longer affected by the phenomenon, digital reconstruction of solarization through a simple and computationally-cheap non-linear transformation remains feasible.

Our contributions can be summarized as follows:
\begin{itemize}
    \item We present a new adversarial attack based on \textit{image solarization}. Despite being conceptually simple, the attack is effective, cheap to compute, and does not risk destroying the global structure of natural images.
    \item We benchmark common image classification models and popular defenses against our attack in both, adaptive and universal settings, and show that improvements against other corruptions do not always transfer to robustness against our attack.
\end{itemize}

\section{Adversarial Solarization}
When considering an input image denoted as $X \in \mathbb{R}^{W\times H\times C}$ with dimensions $W\times H$ and $C$ channels (for example, 3 for RGB inputs), solarization $\Psi_{sol}^{\alpha}:X \rightarrow X^{\alpha}$ can be formally defined as follows:
\begin{align}
    X^{\alpha}_{i,j,k} &= \left\{\begin{array}{ll}
                1 - X_{i,j,k} & X_{i,j,k} \geq \alpha\\
                X_{i,j,k} & \mathit{else.}
                \end{array}\right
                .
\end{align}
Here, $\alpha \in [0, 1]$ serves as a threshold determining the point at which pixel tones undergo inversion. Notably, for $\alpha=0$, the solarized image becomes an inverted version of the original, while $\alpha=1+\epsilon$ maintains the image's unmodified state. To see some examples of this transformation, refer to the examples depicted in Figure \ref{fig:solarization_attack}. Observing the attack, we notice its primary impact on color information. In specific instances (\textit{binder}) it even introduces new edges. Nevertheless, across all cases, the global shape remains consistent. This observation leads us to an intriguing contrast: humans typically rely on shapes for object recognition, whereas models often prioritize texture \cite{geirhos2018imagenettrained}. This presents a compelling scenario with solarization, where label integrity for humans is maintained, but not necessarily for models.

Within the goal to turn solarization into an adversarial attack, the objective is to determine $\alpha$ that satisfies the condition:
\begin{equation}
    \mathcal{F}\big(\Psi_{sol}^{\alpha}(X); \theta\big) \neq \mathcal{F}(X; \theta).
\end{equation}
Ultimately, the sought-after $\alpha$ is chosen to induce a discrepancy in the network's predictions when applied to the perturbed image compared to the original image.

In this context, robust accuracy - this is the accuracy exhibited under the adversarial solarization attack - is defined as:%
\begin{align}
\frac{1}{|\mathcal{D}|} \sum_{(x,y) \in \mathcal{D}} \mathbb{I}\Big(y \in \text{Top-$k$ Pred. of } \mathcal{F}\big( \Psi_{sol}^{\alpha}(X); \theta\big)\Big).
\end{align}
where $\mathbb{I}$ represents the identity function and Top-$k$ predictions refer to the $k$ labels with the highest predicted probability.%
\begin{figure}
    \centering
    \includegraphics[width=0.85\columnwidth]{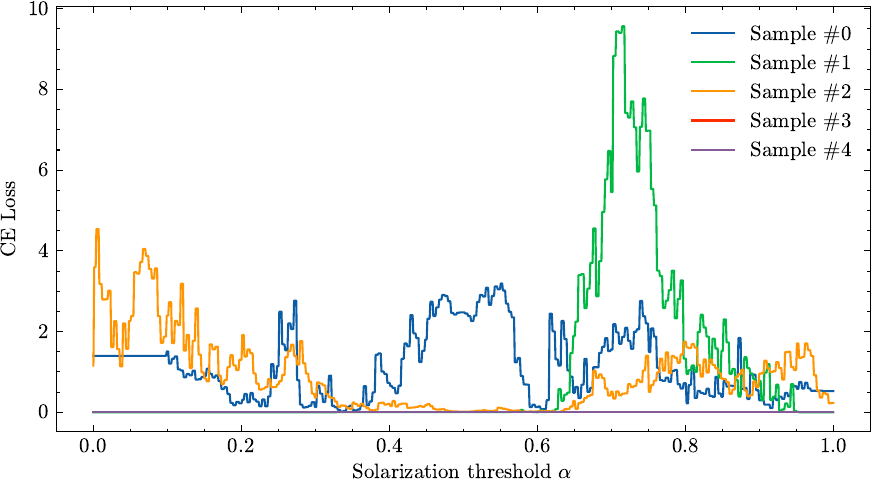}
    \caption{Loss landscape of \textit{ResNet-50} from \textit{timm} \cite{wightman2021resnet} predictions under non-adversarial solarization of 5 ImageNet validation samples.}
    \label{fig:imagenet_loss_landscape}
\end{figure}

Next, we delve into the assessment of solarization's influence on model predictions to derive an optimization method. To this end, we select five samples from the \textit{ImageNet} \cite{imagenet} validation set, measuring cross-entropy loss against \textit{ResNet-50} \cite{resnet} predictions across the full range of possible $\alpha$ values. \Cref{fig:imagenet_loss_landscape} depicts the resulting diverse and intricate loss landscapes. While some samples exhibit minimal impairment irrespective of solarization, others manifest increased loss within specific $\alpha$ ranges. Notably, these landscapes are inherently noisy, sometimes featuring disjoint $\alpha$ ranges with peaks of prediction error. Given these landscapes, gradient-based searches for optimal $\alpha$ values, as utilized in adversarial attacks \cite{SzegedyZSBEGF13}, are less likely to succeed. As a result, we deviate to a greedy random search strategy for $\alpha$ - an approach conceptually simpler and insensitive to the loss landscape due to being gradient-free. To enhance the probability of discovering an adversarial $\alpha$, we iteratively repeat the search until the solarization successfully alters the predicted label for the corresponding sample which can be quantified by any top-\textit{k} accuracy. For the sake of convergence and efficiency, we cap the search at a maximum of $n$ iterations per sample. Consequently, we refer to this attack configuration as \textit{RandSol-Top\{$k$\}-\{$n$\}} in this paper.

Through these formal definitions and assessments, we establish the foundational framework for our exploration of image solarization as an avenue for adversarial attacks.
\section{Experiments}
\subsection{RandSol Attacks}

Our exploration continues by assessing the classification performance of legacy and modern \textit{ImageNet} \cite{imagenet} models. Throughout our experiments, we opt for a fixed value of $n=10$, and $k\in\{1,5\}$ (\textit{RandSol-Top\{1,5\}-10}). This choice of $n$ was made based on empirical findings that balance the trade-off between search time and attack effectiveness. Further increasing $n$ will result in higher attack success rates at an increased search time.

The findings, summarized in \cref{tab:eval}, showcase a trade-off between top-1 and top-5 attacks. Top-1 attacks prove effective at reducing top-1 accuracy but exhibit less effectiveness against top-5 accuracy. Conversely, top-5 attacks lead to higher error rates in top-5 accuracy but show reduced effectiveness in impacting top-1 accuracy. Since the majority of methods primarily emphasize top-1 accuracy \cite{SzegedyZSBEGF13,hendrycks2018benchmarking,mintun2021on,madry2018towards}, we proceed with our investigation, focusing exclusively on the top-1 attack variant.

Depending upon the model under examination, significant differences in robustness arise.
Legacy models exhibit a notable drop in top-1 accuracy, often falling below the $20\%$ mark. In contrast, modern networks demonstrate considerably greater resilience, a distinction that we do not attribute to the network itself but rather to the augmentations used during training. It becomes evident that models relying on \textit{RandAug} \cite{cubuk2020} training, incorporate solarization as an augmentation, and thus increase robustness against the same. This is visible when contrasting the performance of \textit{ResNet-50} trained with conventional \textit{ImageNet} training as in \cite{resnet} versus the upgraded \textit{timm} training \cite{wightman2021resnet} featuring \textit{RandAug}. In these instances, the attack's perturbations can no longer be strictly considered out-of-distribution and unsurprisingly are thus easier to classify. Nonetheless, even these augmented models are significantly affected by the \textit{RandSol-10} attack, leading to top-1 accuracy reductions of up to $27\%$. Overall, \textit{ConvNeXt-B} \cite{convnext} exhibits the most robust performance, surpassing \textit{ViT-B/16} \cite{vit}, which excels in clean data scenarios.
\begin{table}
    \small
    \centering
    \caption{Evaluation of various \textit{ImageNet} models against the \textit{RandSol-10} attack on the top-1/5 accuracy. Gray background indicates models trained on solarization-augmented samples.}
    \label{tab:eval}
    \vskip 0.5em
    \resizebox{\columnwidth}{!}{
    \begin{tabular}{l|rr|rr|rr} %
    \toprule
                         & \multicolumn{2}{c}{\textbf{Clean}}                                          & \multicolumn{2}{c}{\textbf{R.S.-Top1-10}}                                & \multicolumn{2}{c}{\textbf{R.S.-Top5-10}}                                \\ 
    \textbf{Model}       & \multicolumn{1}{c}{Top1} & \multicolumn{1}{c}{Top5} & \multicolumn{1}{c}{Top1} & \multicolumn{1}{c}{Top5} & \multicolumn{1}{c}{Top1} & \multicolumn{1}{c}{Top5} \\ \midrule
    AlexNet \cite{NIPS2012_c399862d}              & 56.52                                            & 79.07                                            & \textbf{3.20}                                             & 30.98                                            & 7.91                                             & \textbf{10.81}                                            \\
    VGG-16bn \cite{simonyan2015a}            & 73.37                                            & 91.52                                            & \textbf{11.70}                                            & 51.44                                            & 22.61                                            & \textbf{29.53}                                            \\
    RepVGG-B1 \cite{ding2021repvgg}           & 78.37                                            & 94.10                                            & \textbf{19.26}                                            & 62.20                                            & 33.84                                            & \textbf{42.64}                                            \\
    DenseNet-121 \cite{Huang_2017_CVPR}         & 74.43                                            & 91.97                                            & \textbf{17.97}                                            & 60.92                                            & 32.12                                            & \textbf{41.05}                                            \\
    ResNet-18  \cite{resnet}          & 69.76                                            & 89.07                                            & \textbf{10.17}                                            & 49.46                                            & 20.74                                            & \textbf{27.07}                                            \\
    ResNet-34 \cite{resnet}           & 75.12                                            & 92.28                                            & \textbf{16.21}                                            & 57.78                                            & 29.09                                            & \textbf{36.72}                                            \\
    ResNet-50 \cite{resnet}           & 76.13                                            & 92.86                                            & \textbf{13.42}                                            & 53.78                                            & 25.12                                            & \textbf{32.27}                                            \\ \midrule
    \rowcolor{Gray}ResNet-50 (timm) \cite{wightman2021resnet}     & 80.11                                            & 94.50                                            & \textbf{64.78}                                           & 91.93                                            & 75.38                                            & \textbf{86.88}                                          \\
    \rowcolor{Gray}ResNet-101 (timm) \cite{wightman2021resnet}   & 81.90                                            & 95.70                                            & \textbf{71.18}                                           & 94.56                                            & 79.48                                            & \textbf{91.49}                                            \\
    \rowcolor{Gray}EfficientNet-B0 (RA) \cite{pmlr-v97-tan19a} & 76.14                                            & 92.99                                            & \textbf{59.83}                                            & 89.81                                            & 70.79                                            & \textbf{84.63}                                            \\ 
    \rowcolor{Gray}ConvNeXt-B \cite{convnext}          & 83.75                                            & 96.70                                            & \textbf{71.46}                                            & 94.91                                            & 80.20                                            & \textbf{91.49}                                            \\
    \rowcolor{Gray}ViT-B/16 \cite{vit}            & 84.40                                            & 97.27                                            & \textbf{65.14}                                            & 93.47                                            & 77.41                                            & \textbf{89.00}\\ \bottomrule                      
    \end{tabular}
    }
\end{table}
\subsection{Defense Strategies}
Having observed the enhancement in robustness against \textit{RandSol-10} by incorporating solarization into training augmentations, our inquiry delves deeper into exploring alternative defense strategies. Specifically, we turn our attention to defenses that do not integrate solarization into training, thereby offering a genuine assessment of out-of-distribution robustness. Exemplarily, we study various methods built around \textit{ResNet-50} and the \textit{ImageNet} dataset. In particular, we test the \textit{NoisyMix} \cite{erichson2022noisymix} training approach, currently positioned as the most potent \textit{ResNet-50} method as per \textit{RobustBench} \cite{robustbench} against \textit{ImageNet-C} \cite{hendrycks2018benchmarking}. Additionally, we assess \textit{PGD}-based \textit{adversarial training} (\textit{AT}) \cite{madry2018towards}, recognized for its efficacy in improving robustness against $\ell_p$-norm attacks, and improving alignment to human vision in terms of shape-bias \cite{Gavrikov_2023_CVPRW}. Furthermore, we incorporate a model \cite{Beyer_2022_CVPR} distilled from \textit{ResNet-152} pretrained on the \textit{ImageNet21k} dataset \cite{ridnik2021imagenetk}, considering the common correlation between data scaling and enhanced robustness. For each of these methods, we report top-1 accuracy evaluation on clean data and against \textit{RandSol-Top1-10}, and an additional examination of average accuracy across all 19 corruptions and severity levels in \textit{ImageNet-C}. This comprehensive reporting provides insights into potential correlations and trade-offs across these three diverse evaluation scenarios.
\begin{table}
    \centering
    \scriptsize
    \caption{Ablation of various \textit{ResNet-50} methods on the top-1 test accuracy on clean \textit{ImageNet}, \textit{ImageNet-C} (avg.), and our \textit{RandSol-10} attack.}
    \label{tab:defenses}
    \vskip 0.5em
        \begin{tabular}{lrrr}
        \toprule
         & \multicolumn{3}{c}{\textbf{Top-1 Test Accuracy} [\%]} \\
        \textbf{Model}                  & Clean          & ImageNet-C & RandSol-10     \\ \midrule
        ResNet-50 \cite{resnet}              & 76.13          & 39.57      & 13.48          \\
        + NoisyMix \cite{erichson2022noisymix}      &   77.05        &  52.35         & 33.33 \\
        + AT ($\ell_\infty, \epsilon=\frac{4}{255}$) \cite{madry2018towards} & 63.87          & 32.39      & 6.67           \\
        + BiT-M Distillation \cite{Beyer_2022_CVPR}    & \textbf{82.81} &       \textbf{ 58.04}    & 25.30          \\ 
        \midrule
        \rowcolor{Gray}+ timm \cite{wightman2021resnet}                 & 80.11          & 48.44      & \textbf{64.77}         \\
        \bottomrule
        \end{tabular}%
\end{table}
The results in Table \ref{tab:defenses}, show intriguing observations. \textit{NoisyMix} substantiates its value by significantly elevating performance beyond the baseline. However, it remains notably vulnerable, especially when contrasted with the resilience exhibited by \textit{RandAug}-based training. Curiously, despite its efficacy in combating $\ell_p$-bound adversarial attacks, adversarial training (\textit{AT}) displays as distinct overfitting tendency to the training attack, leading to a considerable decline in performance across all three tests. Lastly, the distilled model showcases impressive results in both clean accuracy and \textit{ImageNet-C} performance. However, its ability to withstand \textit{RandSol} is not as effective as \textit{NoisyMix's}. This finding effectively underscores the nuanced nature of robustness benchmarks on static datasets, which might not necessarily transfer to other dimensions of robustness.

Across all instances, the \textit{RandSol} attack consistently results in a notable decrease in accuracy, indicating the need for further exploration and refinement of defense strategies.
\begin{figure}
    \centering
    \includegraphics[width=\columnwidth]{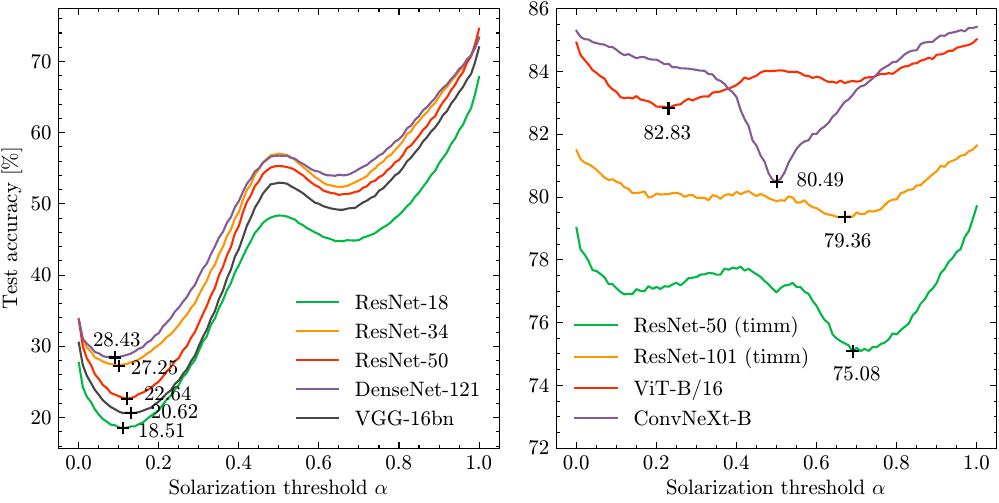}
    \caption{Top-1 accuracy of various models under universal solarization of \textit{ImageNet} validation samples. Please note the different y-scales between the plots.} %
    \label{fig:imagenet_nonadv_sol}
\end{figure}
\subsection{Universal Solarization Attacks}

In the final phase of our experiments, our focus shifts to the exploration of whether universally applicable $\alpha$ values exist, capable of successfully deceiving models. The discovery of such values would be offering a potent black-box attack strategy. It is important to note that, from a conceptual standpoint, this approach inherently holds less potential than \textit{RandSol}, where the adjustment of $\alpha$ per sample enables the maximization of misclassification rates.

Reverting to the same experimental setup as utilized for the \textit{RandSol} method, we now adopt an approach in which all samples are solarized using a constant $\alpha$ value. The range of interest for $\alpha$ spans from 0 to 1, incrementing in steps of 0.01.

The outcomes, as presented in Table \ref{fig:imagenet_nonadv_sol}, once again underscore the differences between models trained via the conventional 90-epoch \textit{ImageNet} training regimen with simple augmentations and those subjected to more sophisticated training schemes like \textit{RandAug}.
For the former set of models, it's noteworthy that a straightforward inversion of images (achieved at $\alpha=0$) yields substantial performance deterioration. These models exhibit a consistent response to universal $\alpha$ values, slightly biased by their overall clean performance. In general, the global minimum for these models typically lies around $\alpha \approx 0.12$.

In stark contrast, modern models demonstrate a resiliance to universal solarization attacks. However, they are not immune to such strategies and still reveal inherent vulnerabilities. Interestingly, the optimal attack parameters tend to correlate with the specific training schemes as the models display a higher degree of variance in trends, often diverging significantly in the positions of their global minima. For instance, \textit{ResNet-50} has a global minimum at $\alpha=0.69$, \textit{ConvNeXt-B} at $\alpha=0.5$, and \textit{ViT-B/16} at $\alpha=0.23$. Yet, models trained under the same regime, such as \textit{timm's ResNet-50/101}, exhibit analogous trends, albeit again influenced by their accuracy levels similarly to legacy models.
\section{Conclusion}
In a step towards more holistic robustness evaluations we present a new adversarial attack based on image solarization which is conceptually simple, yet effective, and preserves global shape information independent of intensity, while being cheap to compute. Our evaluations show that regularly trained models are highly vulnerable to this attack. Methods that increase robustness against other corruptions do not necessarily transfer to this setting. Unsurprisingly, models that use solarization as augmentation during training (e.g., \textit{RandAug}) are more robust but still show an impairment of accuracy. In both cases, model-dependant universal weak spots exist, that may be exploited by malicious actors.

\bibliographystyle{IEEEbib_mod}
\bibliography{main}

\end{document}